\documentclass[letterpaper, 10 pt, conference]{ieeeconf}  

\IEEEoverridecommandlockouts                              

\usepackage{graphicx} 
\usepackage{epsfig} 
\usepackage{hyperref}
\usepackage{times}      
\usepackage{xcolor}
\usepackage{amsmath} 
\usepackage{amssymb}  
\usepackage{cite}
\usepackage{array}
\usepackage{multirow}
\usepackage{amssymb}
\usepackage[caption=false,font=footnotesize]{subfig}

\newcommand{\m}[1]{{PACE}}

\newcommand{\disttau}{\boldsymbol{\tau}_{\mathrm d}}
\newcommand{\estdisttau}{\hat{\boldsymbol{\tau}}_{\mathrm d}}
\newcommand{\dotdisttau}{\dot{\boldsymbol{\tau}}_{\mathrm d}}

\newcommand{\tablebodyfont}{\footnotesize}

\graphicspath{{./}}

\title{\LARGE \bf
\m{}: Policy Augmentation and Compensation With Explicit Disturbance Estimation for Robust Locomotion
}

\author{
Shixuan Han$^{1}$,
Yuyang Wang$^{2}$,
Jingting Jia$^{3}$,
Zeshuai Chen$^{1}$,
Jindou Jia$^{4,\dagger}$,
Kexin Guo$^{5,\dagger}$,
Xiang Yu$^{1}$%
\thanks{$^{1}$School of Automation Science and Electrical Engineering,
Beihang University, Beijing, China;
$^{2}$School of Astronautics, Beihang University, Beijing, China;
$^{3}$Hangzhou International Innovation Institute,
Beihang University, Hangzhou, China;
$^{4}$Nanyang Technological University, Singapore;
$^{5}$School of Aeronautic Science and Engineering,
Beihang University, Beijing, China.
$^{\dagger}$Corresponding author.}%
}

\begin{document}

\maketitle
\thispagestyle{empty}
\pagestyle{empty}

\begin{abstract}
Reinforcement learning has made quadrupedal locomotion remarkably robust. To handle unmodeled dynamics and external disturbances, domain randomization is usually employed to make a policy hedge against every condition it might meet. However, maintaining accurate tracking across varying dynamics and generalizing beyond the sampled range remain challenging. Here we show that explicitly estimating the disturbance and feeding the resulting estimation back into both the policy and the controller can relax this compromise. We introduce Policy Augmentation and Compensation with Explicit disturbance estimation (PACE). The framework lumps payload variations, center-of-mass (CoM) shifts, actuator degradation, and external forces into one joint-space disturbance. A nonlinear observer subsequently recovers this disturbance online without requiring joint acceleration measurements. The disturbance estimation provides feedforward torque compensation as the foundational control pathway and further augments the policy observation to enable disturbance-aware action generation. Considering the learning difficulty of fast-varying contact effects, we further provide an analytical contact estimation approach from joint and inertial measurements, without demanding force sensors. Across eleven simulated dynamics conditions, \m{} reduces velocity-tracking error by 18.1\% to 57.0\% against state-of-the-art baselines, and maintains improved tracking under out-of-distribution (OOD) commands. Notably, in real-world experiments, the deployed real robot carried an 11.5 kg payload, nearly equal to its own mass. It also negotiated unseen 16 cm step-downs and crossed five outdoor terrains. The policy was trained exclusively on flat ground in simulation with at most 3 kg of added mass.
\end{abstract}

\section{INTRODUCTION}

Quadrupedal robots are increasingly considered for industrial inspection, search and rescue, and exploration in unstructured environments~\cite{c1,c2,c3}. Unlike wheeled platforms, they negotiate discontinuous and irregular terrain by actively selecting footholds and coordinating the whole body, which extends the operational range of autonomous systems. In practice, however, this capability depends on the controller's robustness against external disturbances such as payload and pushes, as well as internal model uncertainties~\cite{jia2026physics}.

Model-based approaches have enabled highly dynamic legged locomotion through trajectory optimization~\cite{winkler2018gait}, whole-body control~\cite{c5}, and model predictive control~\cite{c4,c6,c7}. These pipelines are transparent and offer explicit mechanisms for rejecting disturbances. However, the effectiveness of these methods generally depends on sufficiently accurate dynamics models. Constructing such models demands task-specific expertise, careful calibration, and extensive tuning. Moreover, the locomotion performance may deteriorate once the real dynamics drift away from the design assumptions. Payload changes, actuator degradation, and unmodeled contact all produce exactly this kind of drift. Especially, the contact one acts as a step-like disturbance that is not necessarily covered by the guarantees of classical anti-disturbance theories.

Deep reinforcement learning (RL) removes the need for hand-designed gait rules by learning behaviors through interaction. Combined with massively parallel simulation, RL has enabled agile and increasingly capable locomotion~\cite{jia2026physics,c8,c9,c10,c11,c12,c13,c14,c15,c16,c17,c18, jia2026ode}. Robustness to varying dynamics is pursued along three broad lines. The first is domain randomization, which broadens the training distribution so that a single policy covers diverse physical parameters~\cite{c19}. The second infers dynamics variations online as a latent code or an estimated state~\cite{c11,c14,c20,c21}. The third attaches a model-based adaptive mechanism to the learned policy~\cite{c22}.

The first two lines share a structural property. The dynamics variation is primarily accommodated through the policy, so the policy hedges across every condition it might meet. However, maintaining accurate tracking and generalizing beyond the range sampled during training can remain challenging. Neither line produces a quantity in physical units that can act on the joint torques directly. Thus, the correction is only as fast as policy inference. Moreover, domain randomization may introduce trade-offs between condition-specific tracking accuracy and overall performance~\cite{nerual_fly,jia2024feedback,lyu2026adapt}. The third line restores an explicit compensation channel. However, the adaptive mechanism is introduced after the policy has been trained, leaving a mismatch between the dynamics seen during learning and those seen at deployment.

In this work, rather than relying exclusively on learned representations of varying dynamics, we lump disturbances from the nominal model into a single joint-space disturbance. A nonlinear disturbance observer (DO)~\cite{c23,c24,c25,guo2005disturbance} explicitly estimates this disturbance online. The estimation is first applied as torque-level feedforward compensation to counteract its effects at the joint level. Building upon this compensation pathway, the estimation is additionally provided to the policy as an observation, enabling disturbance-aware action generation. Both pathways are incorporated during training and retained at deployment, allowing the policy to be optimized against dynamics that are already partially corrected. We call the resulting framework \textbf{\m{}}, i.e., \textbf{P}olicy \textbf{A}ugmentation and \textbf{C}ompensation with \textbf{E}xplicit disturbance estimation.

\begin{figure*}[!t]
    \centering
    \setlength{\abovecaptionskip}{0pt}
    \includegraphics[width=\textwidth]{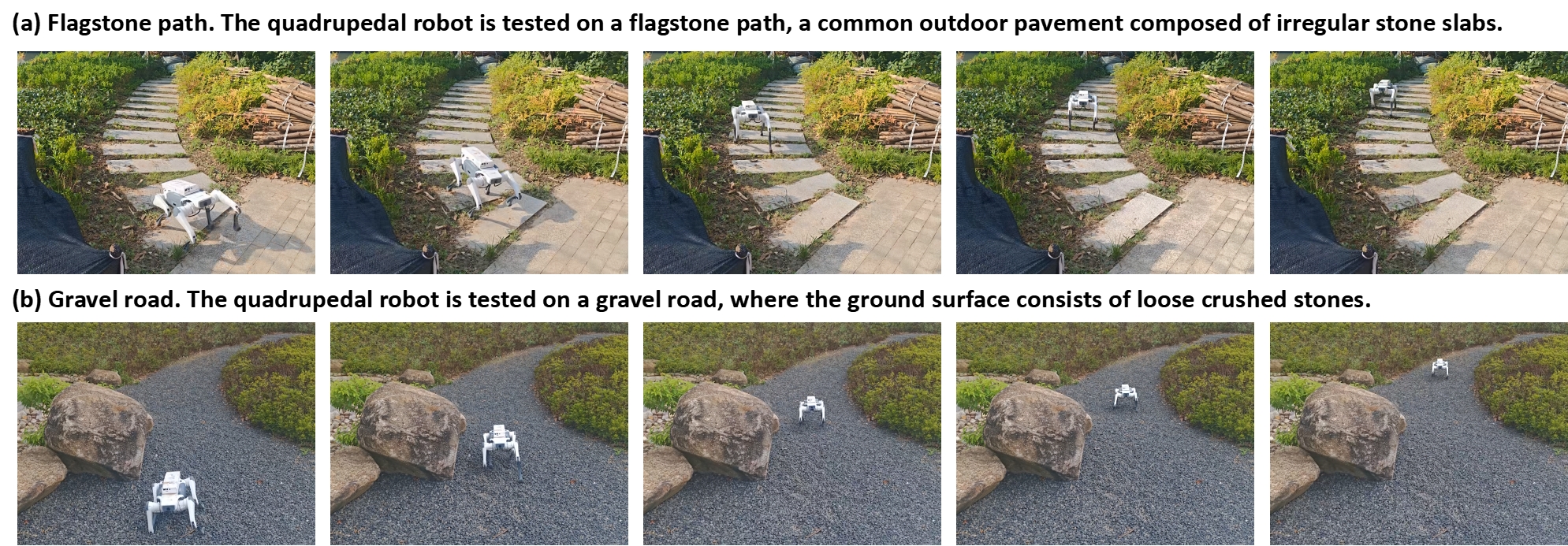}
    \caption{Representative locomotion sequences on unseen
    real-world terrains after training exclusively on flat terrain in simulation.
    (a) Traversal of a flagstone path composed of irregular
    stone slabs.
    (b) Traversal of a gravel road covered with loose stones.
    Successive frames illustrate the robot's progression
    across the two terrain types.}
    \label{fig:terrain_sequences}
\end{figure*}

Furthermore, deploying PACE on hardware requires the contact-force term of the robot dynamics. However, many quadrupedal platforms lack dedicated foot-force sensing or suffer from noisy and unreliable force measurements under repeated foot--ground impacts~\cite{kang2023footforce,zhang2020footforce}. We therefore reconstruct contact forces from joint states and inertial measurements alone, which keeps the observer structurally identical in simulation and on the robot.  Validated in MuJoCo, the reconstruction tracks the simulator reference with a $2.7~\mathrm{N}$ vertical RMSE. More importantly, substituting the reconstructed forces for the simulator contact forces yields a mean joint-wise RMSE of only $0.054~\mathrm{N\cdot m}$ between the corresponding disturbance estimates. This close agreement supports the
use of reconstructed contact forces in the observer during real-world deployment.

PACE is evaluated in simulation under payload, CoM, actuator, and external-force variations, and on a real-world DeepRobotics Lite3 under large payloads, unseen height discontinuities, and five outdoor terrains. Across eleven dynamics-variation conditions, PACE lowers velocity-tracking error by $57.0\%$ and $18.1\%$ relative to BASELINE~\cite{deep} and HIMLOCO~\cite{c21}, respectively. Under an $80~\mathrm{N}$ lateral push, PACE further shortens the recovery time by $64.7\%$ and $76.0\%$ relative to the same two methods, respectively. The advantage persists outside the training distribution. Under the unseen $2~\mathrm{m/s}$ command, PACE maintains stable locomotion while BASELINE initially falls repeatedly. On hardware, PACE carries an $11.5~\mathrm{kg}$ payload, clears $16~\mathrm{cm}$ step-downs, and traverses five unseen outdoor terrains after flat-ground training alone. Representative traversals of a flagstone path and gravel road are shown in Fig.~\ref{fig:terrain_sequences}.

To recap, the main contributions are summarized as:
\begin{itemize}
    \item Heterogeneous model uncertainties and external effects are lumped into a single joint-space disturbance and estimated online by an acceleration-free nonlinear observer. Unlike
    post-training adaptation, the observer remains in the closed loop throughout training and deployment.
    \item Torque-level feedforward compensation forms the foundational enhancement pathway of PACE. Building on this pathway, explicit disturbance estimation further augments the policy observations to enable disturbance-aware action generation.
    \item A sensorless foot-force estimator is developed to reconstruct the contact-force term required by the observer from onboard measurements, preserving the observer structure during real-world deployment.
\end{itemize}

\section{Related Work}
\label{sec:related}

\subsection{Model-Based Locomotion Control}

Model-based approaches have played a central role in dynamic quadrupedal locomotion. Convex model predictive control enables dynamic gaits on the MIT Cheetah 3~\cite{c4}, and combining whole-body impulse control with model predictive control extends this to highly dynamic motions~\cite{c5}. Real-time constrained nonlinear model predictive control on $\mathrm{SO}(3)$ handles the orientation dynamics directly~\cite{c6}, while a representation-free formulation avoids the singularities introduced by attitude parameterization~\cite{c7}. These methods are transparent and physically grounded, but their performance depends on model fidelity and demands careful calibration and tuning. PACE instead uses a learned policy as the primary controller, with a model-based DO acting only as an auxiliary module that estimates and compensates for the residual dynamics.

\subsection{Learning-based Policies for Generalization}

\begin{figure*}[t]
    \centering
    \setlength{\abovecaptionskip}{0pt}
    \includegraphics[width=\textwidth]{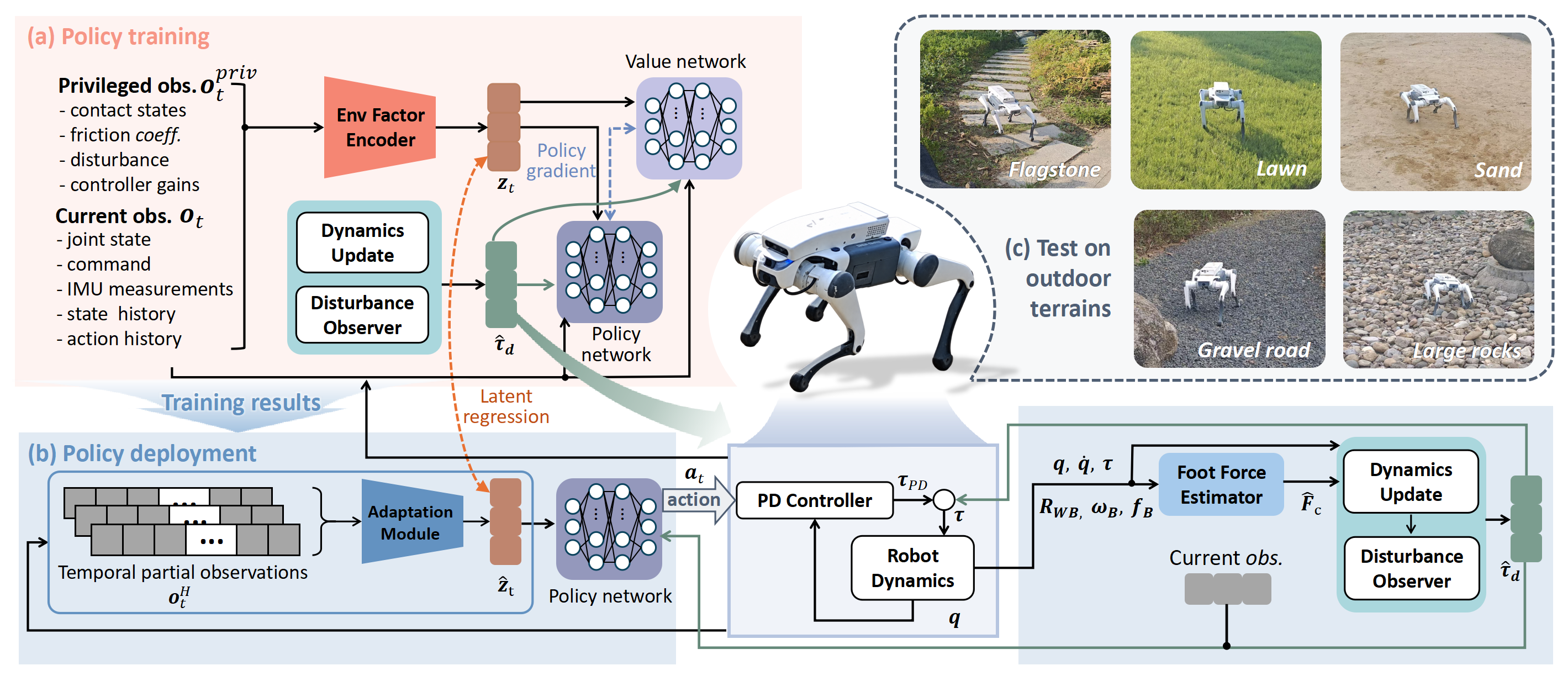}
    \caption{Overview of the proposed PACE framework. 
    (a) During training, privileged observations are encoded into the latent representation $\mathbf{z}_t$, while the explicit disturbance estimation is used for torque-level feedforward compensation and further provided to the policy as an additional observation. 
    (b) During deployment, the adaptation module infers $\hat{\mathbf{z}}_t$ from temporal partial observations, and the sensorless foot-force estimator reconstructs the contact-force input required by the DO, preserving the same disturbance-aware closed-loop structure without privileged information or foot-force sensors. 
    (c) The policy, trained exclusively on flat terrain in simulation, is evaluated zero-shot on five unseen outdoor terrains.}
    \label{fig:framework}
    \vspace{-1pt}
\end{figure*}

Learning-based locomotion has enabled increasingly agile and versatile quadrupedal behaviors~\cite{c8,c9,c10,c11,c12,c13,c14,c15,c16,c17,c18}. To improve robustness to dynamics variations, dynamics randomization exposes the policy to diverse physical parameters during training~\cite{c19}. Rapid Motor Adaptation (RMA) infers latent environmental factors from proprioceptive histories~\cite{c11}. Concurrent training jointly optimizes a locomotion policy and a state estimator~\cite{c20}. DreamWaQ combines explicit state estimation with implicit environmental context~\cite{c14}. HIMLOCO characterizes environmental variations through the internal response of the robot~\cite{c21}. These approaches improve generalization, but limitations remain. Dynamics randomization may sacrifice condition-specific tracking accuracy for average performance over the sampled conditions~\cite{nerual_fly,jia2024feedback,lyu2026adapt}. Learned latent representations often remain abstract and lack direct physical interpretability. PACE complements learned latent representations with an explicitly estimated joint-space disturbance in physical units, providing a physically interpretable measure of dynamics discrepancies. Moreover, the contact force enters this estimate analytically rather than being inferred solely through a learned latent representation.

\subsection{Combining Learning With Model-Based Adaptation}

Recent work has explored combining learned locomotion policies with model-based adaptation to improve robustness. RL2AC~\cite{c22} augments a pretrained policy with a plug-and-play online adaptive controller, compensating for dynamics discrepancies encountered during deployment. A related line inserts a physics-based filter beneath the learned policy~\cite{jia2026physics}, where the estimated residual is applied as a low-level torque correction while the policy input is left unchanged. Unlike the post-training adaptation in RL2AC, PACE incorporates the DO into the closed loop from the beginning of policy training. Beyond foundational torque-level feedforward compensation, the disturbance estimation further augments the policy observation, enabling disturbance-aware action generation.  Moreover, the contact force is treated as an explicit term of the dynamics rather than being absorbed into the residual like previous works.

\section{Preliminaries}

\subsection{Reinforcement Learning}

We cast quadrupedal locomotion as an infinite-horizon partially observable \textit{Markov} decision process (POMDP) $\mathcal{M}=(\mathcal{S},\mathcal{O},\mathcal{A},d_0,p,r,\gamma)$, with continuous state $\mathbf{s}\in\mathcal{S}$, observation $\mathbf{o}\in\mathcal{O}$, and action $\mathbf{a}\in\mathcal{A}$. An episode starts from $\mathbf{s}_0\sim d_0$ and evolves under the transition density $p(\mathbf{s}_{t+1}\,|\,\mathbf{s}_t,\mathbf{a}_t)$. Every transition returns a reward $r:\mathcal{S}\times\mathcal{A}\rightarrow\mathbb{R}$, and the policy $\pi$ is optimized to maximize the expected discounted return
\begin{equation}
J(\pi)=\mathbb{E}\left[\sum_{t=0}^{\infty}\gamma^{t}\,r(\mathbf{s}_t,\mathbf{a}_t)\right],
\end{equation}
where $\gamma\in[0,1)$ is the discount factor.

The robot has access only to $\mathbf{o}_t\in\mathbb{R}^{117}$, which excludes the physical parameters governing its own dynamics. Two quantities are therefore introduced. The observation history over $H=40$ consecutive steps, $\mathbf{o}_t^{H}=[\mathbf{o}_t^{\mathbf{T}},\mathbf{o}_{t-1}^{\mathbf{T}},\ldots,\mathbf{o}_{t-H+1}^{\mathbf{T}}]^{\mathbf{T}}\in\mathbb{R}^{4680}$, carries temporal proprioceptive information about how the robot responds to its own actions. The latent vector $\mathbf{z}_t\in\mathbb{R}^{18}$ compactly encodes the environmental and dynamical factors that are accessible in simulation but not on hardware. During deployment, $\mathbf{z}_t$ is replaced by the estimate $\hat{\mathbf{z}}_t\in\mathbb{R}^{18}$ that the adaptation module infers from $\mathbf{o}_t^{H}$.

\subsection{Quadrupedal Dynamics and Low-level Controller}

Let $\mathbf{q},\dot{\mathbf{q}},\ddot{\mathbf{q}}\in\mathbb{R}^{12}$ denote the actual joint position,
velocity, and acceleration vectors of the quadrupedal robot, respectively.  We consider the following joint-level representation of the robot dynamics:
\begin{equation}
\label{eq:robot_dynamics}
\mathbf{M}(\mathbf{q})\ddot{\mathbf{q}}
+
\mathbf{C}(\mathbf{q},\dot{\mathbf{q}})\dot{\mathbf{q}}
+
\mathbf{G}(\mathbf{q})
-
\mathbf{J}^{\mathbf{T}}\mathbf{F}_{\mathrm{c}}
=
\boldsymbol{\tau},
\end{equation}
where $\mathbf{M}(\mathbf{q})\in\mathbb{R}^{12\times12}$ is the inertia matrix,
$\mathbf{C}(\mathbf{q},\dot{\mathbf{q}})\in\mathbb{R}^{12\times12}$ represents the \textit{Coriolis} and centrifugal
matrix, $\mathbf{G}(\mathbf{q})\in\mathbb{R}^{12}$ denotes the gravity term, $\mathbf{J}\in\mathbb{R}^{12\times12}$ is the contact \textit{Jacobian}
matrix, $\mathbf{F}_{\mathrm{c}}\in\mathbb{R}^{12}$ is the foot contact-force vector, and $\boldsymbol{\tau}\in\mathbb{R}^{12}$ denotes
the joint torque vector.

In RL-based quadrupedal locomotion, the action generated by the policy is
commonly converted into joint torques through a proportional--derivative (PD) controller. The
resulting control torque $\boldsymbol{\tau}_{\mathrm{PD}}\in\mathbb{R}^{12}$ is given by
\begin{equation}
\label{eq:pd_control}
\boldsymbol{\tau}_{\mathrm{PD}}
=
\mathbf{K}_{\mathrm{p}}\left(\mathbf{q}_{\mathrm{des}}-\mathbf{q}\right)
-
\mathbf{K}_{\mathrm{d}}\dot{\mathbf{q}},
\end{equation}
where
$\mathbf{K}_{\mathrm{p}},\mathbf{K}_{\mathrm{d}}\in\mathbb{R}^{12\times12}$
denote the proportional and derivative gain matrices, respectively, and
$\mathbf{q}_{\mathrm{des}}\in\mathbb{R}^{12}$
denotes the desired joint-position command derived from the learned policy.

\section{Method}

\subsection{Overview of the Framework}

The overall architecture of PACE is illustrated in Fig.~\ref{fig:framework}. The learned policy serves as the primary locomotion controller, whose action is converted into desired joint positions and tracked by a joint-level PD controller. In parallel, a nonlinear DO estimates the lumped joint-space disturbance from the nominal robot dynamics and contact information. During simulation, the contact force is directly available from the simulator, whereas during real-world deployment it is reconstructed from onboard joint and inertial measurements by the sensorless foot-force estimator. 

The resulting disturbance estimation is utilized through two complementary paths: it is applied as torque-level feedforward compensation and further provided to the policy as an additional observation. Moreover, the observer remains in the closed loop throughout both training and deployment. Unlike approaches that modify the control loop only after policy training, PACE optimizes the policy under partially corrected dynamics.

\subsection{Disturbance Observer}
\label{sec:disturbance_observer}

We first introduce the employed analytic DO. Payload variations, CoM shifts, actuator degradation, external forces, and model mismatch can all introduce joint-space dynamics that deviate from the nominal robot model.
Instead of identifying these effects individually, we aggregate their equivalent influence into a
lumped disturbance $\disttau\in\mathbb{R}^{12}$. For the DO, the inertia matrix is approximated by a constant diagonal nominal matrix $\mathbf{M}\in\mathbb{R}^{12\times12}$. This representation reduces the online computational burden and avoids relying on an exact configuration-dependent inertia model. Instead, the neglected coupling effects, configuration dependence, and parameter mismatch are absorbed into the lumped disturbance $\disttau$. This choice trades modeling fidelity for practicality, and the real-robot results confirm that the simplified nominal model remains sufficient. The robot dynamics can therefore be written as

\begin{equation}
\label{eq:disturbed_dynamics}
\mathbf{M}\ddot{\mathbf{q}}
+
\mathbf{C}(\mathbf{q},\dot{\mathbf{q}})\dot{\mathbf{q}}
+
\mathbf{G}(\mathbf{q})
-
\mathbf{J}^{\mathbf{T}}\mathbf{F}_{\mathrm{c}}
=
\boldsymbol{\tau}
+
\disttau.
\end{equation}

Following the nonlinear DO formulation
in~\cite{c23,c25,guo2005disturbance}, we construct an acceleration-free
observer for the lumped disturbance
$\boldsymbol{\tau}_{\mathrm{d}} \in \mathbb{R}^{12}$ as
\begin{equation}
\label{eq:eta_dynamics}
\begin{aligned}
\dot{\boldsymbol{\eta}}
=
-L\boldsymbol{\eta}
+
L\big(
&
\mathbf{C}(\mathbf{q},\dot{\mathbf{q}})\dot{\mathbf{q}}
+
\mathbf{G}(\mathbf{q})
\\
&
-
\mathbf{J}^{\mathbf{T}}\mathbf{F}_{\mathrm{c}}
-
\boldsymbol{\tau}
-
\boldsymbol{\rho}(\dot{\mathbf{q}})
\big),
\end{aligned}
\end{equation}
with the disturbance estimate reconstructed by
\begin{equation}
\hat{\boldsymbol{\tau}}_{\mathrm{d}}
=
\boldsymbol{\eta}
+
\boldsymbol{\rho}(\dot{\mathbf{q}}),
\label{eq:do_estimate}
\end{equation}
where $\boldsymbol{\eta}\in\mathbb{R}^{12}$ is the auxiliary
variable, $L>0$ is a constant scalar observer gain, and
$\boldsymbol{\rho}(\dot{\mathbf{q}})
=L\mathbf{M}\dot{\mathbf{q}}$.
Since $\mathbf{M}$ and $L$ are constant, the auxiliary-variable
formulation eliminates the explicit dependence on joint
acceleration. Under the nominal dynamics, the observer
satisfies
$\dot{\hat{\boldsymbol{\tau}}}_{\mathrm{d}}
=L(\boldsymbol{\tau}_{\mathrm{d}}-\hat{\boldsymbol{\tau}}_{\mathrm{d}})$.

To analyze the estimation behavior, we define the disturbance estimation error $\mathbf{e}_{\mathrm{d}}\in\mathbb{R}^{12}$ as
\begin{equation}
\label{eq:do_error}
\mathbf{e}_{\mathrm{d}}
=
\disttau
-
\estdisttau.
\end{equation}

Differentiating the estimation error and using the observer dynamics yields
\begin{equation}
\label{eq:do_error_dynamics}
\dot{\mathbf{e}}_{\mathrm{d}}
=
\dotdisttau
-
L\mathbf{e}_{\mathrm{d}}.
\end{equation}

Assume that the lumped disturbance is continuously differentiable with a bounded variation rate,
\begin{equation}
    \left\|\dot{\boldsymbol{\tau}}_{\mathrm{d}}(t)\right\|_2
    \leq \bar{d},
\end{equation}
where $\bar{d}<\infty$. This assumption is reasonable. Payload variations, CoM shifts, actuator degradation, friction, and parameter mismatch are either constant or evolve on a time scale far slower than the observer dynamics. High-frequency components need not be estimated, as the robot's own inertia filters them out before they influence the body. The genuinely fast component, the foot contact force, is not absorbed into $\disttau$ at all. It is retained as the explicit term $\mathbf{J}^{\mathbf{T}}\mathbf{F}_{\mathrm{c}}$ in Eq.~\eqref{eq:disturbed_dynamics} and reconstructed analytically in Section~\ref{sec:foot_force}. 

Multiplying both sides of Eq.~\eqref{eq:do_error_dynamics} by the integrating factor $e^{Lt}$ and integrating from $0$ to $t$ gives
\begin{equation}
    \mathbf{e}_{\mathrm{d}}(t)
    =
    e^{-Lt}\mathbf{e}_{\mathrm{d}}(0)
    +
    \int_{0}^{t}
    e^{-L(t-s)}
    \dot{\boldsymbol{\tau}}_{\mathrm{d}}(s)
    \,\mathrm{d}s.
\end{equation}
Taking the \textit{Euclidean} norm and using
$\|\dot{\boldsymbol{\tau}}_{\mathrm{d}}(s)\|_2\leq\bar{d}$ yields
\begin{equation}
\begin{aligned}
    \left\|\mathbf{e}_{\mathrm{d}}(t)\right\|_2
    &\leq
    e^{-Lt}\left\|\mathbf{e}_{\mathrm{d}}(0)\right\|_2 +
    \bar{d}\int_{0}^{t}e^{-L(t-s)}\,\mathrm{d}s \\
    &=
    e^{-Lt}\left\|\mathbf{e}_{\mathrm{d}}(0)\right\|_2
    +
    \frac{\bar{d}}{L}
    \left(1-e^{-Lt}\right).
\end{aligned}
\end{equation}
Therefore,
\begin{equation}
    \limsup_{t\rightarrow\infty}
    \left\|\mathbf{e}_{\mathrm{d}}(t)\right\|_2
    \leq
    \frac{\bar{d}}{L},
\end{equation}
which shows that the disturbance estimation error is uniformly ultimately bounded. For an approximately constant disturbance,
$\dot{\boldsymbol{\tau}}_{\mathrm{d}}\approx\mathbf{0}$, the ultimate bound reduces to zero, and the estimation error converges exponentially to zero. In this work, we set $L=1$ across all experiments.

The disturbance estimation is introduced into the original PD controller as a feedforward compensation term:
\begin{equation}
\label{eq:do_compensation}
\boldsymbol{\tau}
=
\underbrace{
\mathbf{K}_{\mathrm{p}}
\left(\mathbf{q}_{\mathrm{des}}-\mathbf{q}\right)
-
\mathbf{K}_{\mathrm{d}}\dot{\mathbf{q}}
}_{\text{PD control}}
-
\underbrace{
\estdisttau
}_{\text{compensation}}.
\end{equation}

Substituting Eq.~\eqref{eq:do_compensation} into Eq.~\eqref{eq:disturbed_dynamics} yields
\begin{equation}
\label{eq:compensated_dynamics}
\begin{aligned}
\mathbf{M}\ddot{\mathbf{q}}
+
\mathbf{C}(\mathbf{q},\dot{\mathbf{q}})\dot{\mathbf{q}}
+
\mathbf{G}(\mathbf{q})
-
\mathbf{J}^{\mathbf{T}}\mathbf{F}_{\mathrm{c}}
=
&
\mathbf{K}_{\mathrm{p}}
\left(
\mathbf{q}_{\mathrm{des}}-\mathbf{q}
\right)
-
\mathbf{K}_{\mathrm{d}}\dot{\mathbf{q}}
\\
&
+
\disttau
-
\estdisttau .
\end{aligned}
\end{equation}

As $\estdisttau$ approaches $\disttau$, the residual disturbance is attenuated, reducing its influence on the joint dynamics. This feedforward compensation constitutes the foundational enhancement pathway of PACE. As shown in Fig.~\ref{fig:framework}, the disturbance estimation is further provided to the policy as an additional observation, with the corresponding policy architecture detailed in Section~\ref{sec:rl_framework}.

As shown in Eqs.~\eqref{eq:eta_dynamics} and~\eqref{eq:do_estimate}, the DO requires the contact-force term $\mathbf{J}^{\mathbf{T}}\mathbf{F}_{\mathrm{c}}$. Since this term is unavailable from dedicated sensors on the real robot, a sensorless foot-force estimator is introduced next.

\begin{figure}[t]
    \centering
    \setlength{\abovecaptionskip}{0pt}
    \includegraphics[width=\columnwidth]{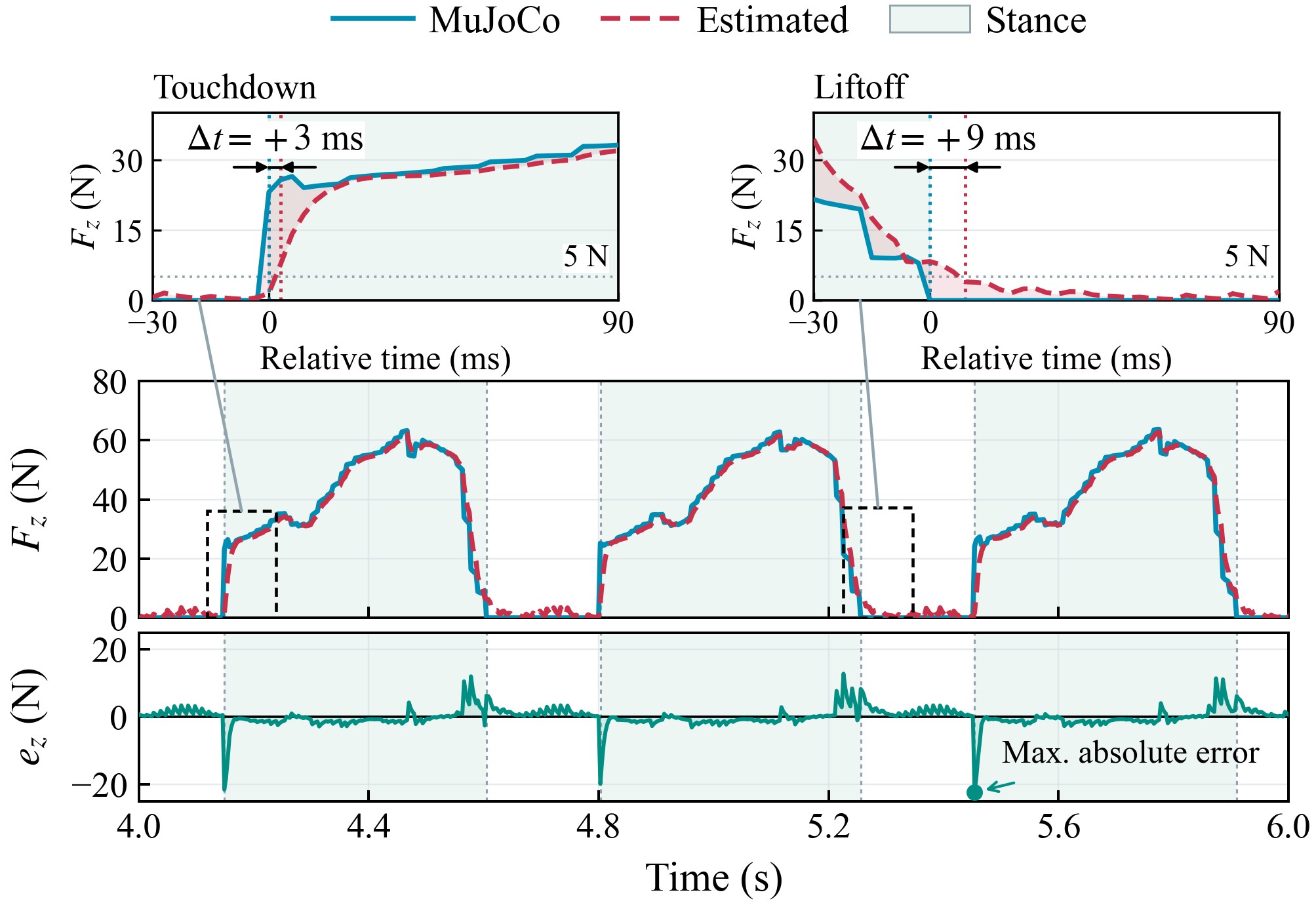}
    \caption{Vertical foot-contact force estimation for the front-left leg
    over $4$--$6\,\mathrm{s}$. The main force plot compares the estimate
    with the MuJoCo reference, and shaded regions indicate reference
    stance intervals. The lower plot shows the signed estimation error,
    with a marker highlighting the maximum absolute error within the
    displayed interval. The two upper insets show enlarged views of
    the regions in the main force plot: touchdown on the left
    and liftoff from the following gait cycle on the right. In these insets, $\Delta t$ denotes the estimation delay.}
\label{fig:foot_force_validation}
    \label{fig:foot_force_validation}
\end{figure}

\subsection{Sensorless Foot-Force Estimator}
\label{sec:foot_force}

The DO requires the contact-force term $\mathbf{J}^{\mathbf{T}}\mathbf{F}_{\mathrm{c}}$, whereas most quadrupedal platforms, including the Lite3 used in this work, either carry no dedicated foot-force sensors or provide readings too noisy for direct use. We therefore develop a sensorless foot-force estimator
using only onboard measurements. At control step $k$, the estimator receives
$\{\mathbf{q}_{k},\dot{\mathbf{q}}_{k},\boldsymbol{\tau}_{\mathrm{m},k},
\mathbf{R}_{\mathrm{WB},k},\boldsymbol{\omega}_{\mathrm{B},k},\mathbf{f}_{\mathrm{B},k}\}$,
where $k$ denotes the discrete control step, $\mathbf{q}$ and $\dot{\mathbf{q}}$ are the joint
position and velocity, $\boldsymbol{\tau}_{\mathrm{m}}\in\mathbb{R}^{12}$ denotes the measured actuator-side joint torque, and
$\mathbf{R}_{\mathrm{WB}}\in SO(3)$ maps vectors from the body frame to the world frame. The inertial measurement unit (IMU) measurements
$\boldsymbol{\omega}_{\mathrm{B}}\in\mathbb{R}^{3}$ and $\mathbf{f}_{\mathrm{B}}\in\mathbb{R}^{3}$ denote the body angular velocity
and specific force, respectively. Throughout this subsection, the superscript $\mathrm{W}$ denotes quantities expressed in the world frame. The estimator reconstructs the three-dimensional contact force $\hat{\mathbf{F}}^{\mathrm{W}}\in\mathbb{R}^{3}$ for each foot, and the four estimated forces are subsequently stacked to form the contact-force input to the DO, denoted by $\hat{\mathbf{F}}_{\mathrm{c}}\in\mathbb{R}^{12}$.

The onboard measurements are preprocessed to obtain the joint
acceleration and the floating-base motion quantities required by inverse
dynamics. The finite-difference, filtering, and frame-transformation
procedures are detailed in the \hyperref[app:preprocessing]{Appendix}. 
For each leg $\ell$, the resulting joint acceleration and floating-base motion quantities, together with the joint states and known inertial parameters, are propagated through a recursive \textit{Newton-Euler} formulation without applying external foot-contact forces, yielding the model-based joint torque $\boldsymbol{\tau}_{\mathrm{\ell}}\in\mathbb{R}^{3}$. The corresponding measured actuator-side joint torque is denoted by
$\boldsymbol{\tau}_{\mathrm{m,\ell}}\in\mathbb{R}^{3}$. The joint-torque residual then satisfies

\begin{equation}
\label{eq:ffe_residual}
\boldsymbol{\tau}_{\mathrm{r,\ell}}
=
\boldsymbol{\tau}_{\mathrm{\ell}}
-
\boldsymbol{\tau}_{\mathrm{m,\ell}}
\approx
(\mathbf{J}_{\mathrm{\ell}}^{\mathrm{W}})^{\mathbf{T}}\mathbf{F}_{\mathrm{\ell}}^{\mathrm{W}},
\end{equation}
where
$\boldsymbol{\tau}_{\mathrm{r,\ell}}\in\mathbb{R}^{3}$,
$\mathbf{J}_{\mathrm{\ell}}^{W}\in\mathbb{R}^{3\times3}$,
and $\mathbf{F}_{\mathrm{\ell}}^{W}\in\mathbb{R}^{3}$
denote the joint-torque residual, translational foot Jacobian, and contact
force exerted by the environment of leg $\ell$, respectively. For simplicity, the leg index is omitted
in the following derivation.


To robustly recover the contact force, define
\begin{equation}
\label{eq:ffe_svd}
\mathbf{A}
=
(\mathbf{J}^{\mathrm{W}})^{\mathbf{T}}
=
\mathbf{U}
\operatorname{diag}(\sigma_{1},\sigma_{2},\sigma_{3})
\mathbf{V}^{\mathbf{T}},
\end{equation}
where $\mathbf{A}\in\mathbb{R}^{3\times3}$ maps the three-dimensional foot force to the
corresponding three joint torques, $\mathbf{U}\in\mathbb{R}^{3\times3}$ and $\mathbf{V}\in\mathbb{R}^{3\times3}$ are orthogonal matrices, and
$\sigma_i$ is the $i$th singular value of $\mathbf{A}$. A damped singular value decomposition (SVD)
inverse is then employed to obtain the raw force estimate $\hat{\mathbf{F}}_{\mathrm{raw}}^{\mathrm{W}}\in\mathbb{R}^{3}$, 
\begin{equation}
\label{eq:ffe_force}
\hat{\mathbf{F}}_{\mathrm{raw}}^{\mathrm{W}}
=
\mathbf{V}
\operatorname{diag}
\left(
\frac{\sigma_i}{\sigma_i^{2}+\lambda^{2}}
\right)
\mathbf{U}^{\mathbf{T}}
\boldsymbol{\tau}_{\mathrm{r}},
\end{equation}
where $\lambda$ denotes the damping coefficient introduced to improve numerical robustness near
singular configurations.

The raw estimate is further projected onto a physically feasible contact-force set by enforcing
unilateral contact, an upper bound on the normal force, and a friction-cone constraint. For
$\mathbf{F}=[\mathbf{F}_{\mathrm{t}}^{\mathbf{T}},F_z]^{\mathbf{T}}\in\mathbb{R}^{3}$,
$\mathbf{F}_{\mathrm{t}}=[F_x,F_y]^{\mathbf{T}}\in\mathbb{R}^{2}$ denotes the tangential force and $F_z$ the normal
force. The projection operator $\mathcal{P}_{\mu}(\cdot)$ maps non-finite estimates or forces with
$F_z\leq0$ to zero and limits the normal force by $F_{z,\max}$. If
$\lVert\mathbf{F}_{\mathrm{t}}\rVert_2>\mu F_z$, the tangential component is projected as
\begin{equation}
\label{eq:ffe_friction_projection}
\mathbf{F}_{\mathrm{t}}
\leftarrow
\frac{\mu F_z}
{\lVert\mathbf{F}_{\mathrm{t}}\rVert_2}
\mathbf{F}_{\mathrm{t}},
\end{equation}
where $\mu$ denotes the friction coefficient. When $\|\mathbf{F}_{\mathrm{t}}\|_2=0$, the scaling operation is skipped to avoid division by zero. The final force estimate is obtained as
\begin{equation}
\label{eq:ffe_filter}
\hat{\mathbf{F}}_{k}^{\mathrm{W}}
=
\mathcal{P}_{\mu}
\left[
(1-\alpha_{\mathrm{F}})\hat{\mathbf{F}}_{k-1}^{\mathrm{W}}
+
\alpha_{\mathrm{F}}
\mathcal{P}_{\mu}
\left(
\hat{\mathbf{F}}_{\mathrm{raw},k}^{\mathrm{W}}
\right)
\right],
\end{equation}
where $\alpha_{\mathrm{F}}$ denotes the output low-pass filtering coefficient. The estimated forces of the four feet are finally stacked as
\begin{equation}
\label{eq:ffe_stacked_force}
\hat{\mathbf{F}}_{\mathrm{c}}
=
\begin{bmatrix}
(\hat{\mathbf{F}}_{\mathrm{FL}}^{\mathrm{W}})^{\mathbf{T}},
(\hat{\mathbf{F}}_{\mathrm{FR}}^{\mathrm{W}})^{\mathbf{T}},
(\hat{\mathbf{F}}_{\mathrm{HL}}^{\mathrm{W}})^{\mathbf{T}},
(\hat{\mathbf{F}}_{\mathrm{HR}}^{\mathrm{W}})^{\mathbf{T}}
\end{bmatrix}^{\mathbf{T}}
\in\mathbb{R}^{12},
\end{equation}
which is supplied to the DO in place of the directly measured contact-force vector $\mathbf{F}_{\mathrm{c}}$ during real-world deployment. In our
implementation, $\alpha_{\mathrm{F}}=0.35$, $\lambda=0.01$, $\mu=1.0$,
and $F_{z,\max}=300~\mathrm{N}$. Although IMU noise can propagate into the reconstructed contact force, the measured inertial signals are filtered before inverse dynamics, while damped inversion and output filtering further suppress noise amplification. As shown in the subsequent real-world experiments, the remaining sensor noise does not prevent practical deployment under the evaluated conditions.

\begin{figure}[t]
    \centering
    \setlength{\abovecaptionskip}{0pt}
    \includegraphics[width=\columnwidth]
    {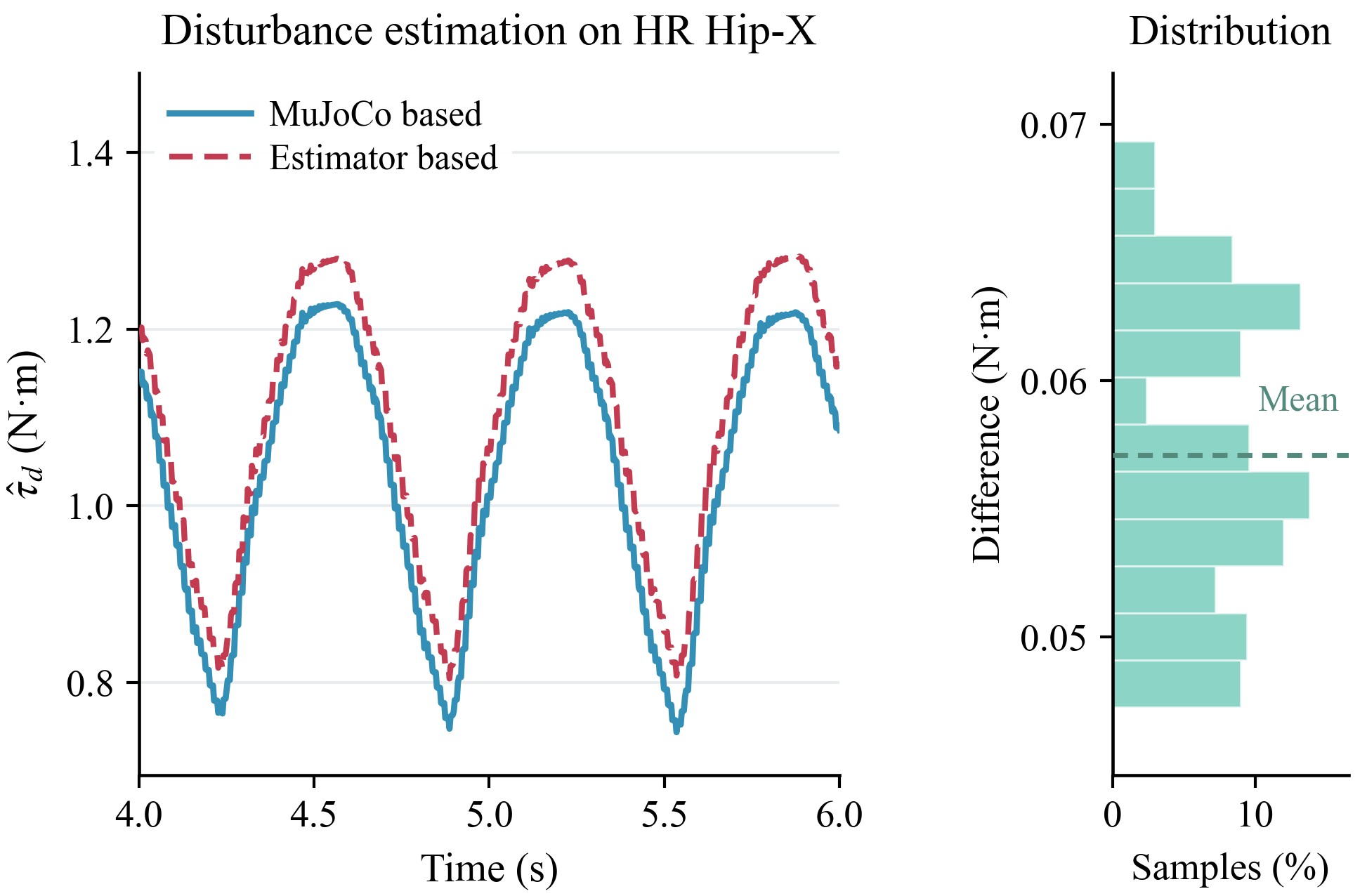}
    \caption{Effect of contact-force substitution on the DO output at the
    HR Hip-X joint. Left: observer outputs obtained using MuJoCo contact
    forces and estimated contact forces over $4$--$6\,\mathrm{s}$.
    Right: distribution of their signed difference using the same samples as in the left plot.
    Horizontal bar lengths indicate the percentage of samples in each
    difference bin, and the dashed line marks the mean difference.}
\label{fig:disturbance_force_curves}
    \label{fig:disturbance_force_curves}
    \label{fig:disturbance_force_curves}
\end{figure}

Validation of the sensorless foot-force estimator is conducted in
MuJoCo in two stages. Force reconstruction is first assessed against
the simulator-reported contact forces. As a representative example,
Fig.~\ref{fig:foot_force_validation} shows the vertical contact force
of the front-left leg. Using the reference force, stance is defined by
$F_z>5\,\mathrm{N}$. The annotated
$\Delta t$ denotes the difference
between the first sampled threshold crossings of the estimated and
reference signals, evaluated on the upward transition for touchdown
and the downward transition for liftoff. At the selected events, the
estimated force exhibits delays of $3\,\mathrm{ms}$ and
$9\,\mathrm{ms}$ for touchdown and liftoff, respectively, indicating
close temporal agreement with the reference contact transitions.
During stance, the estimated force closely follows the reference
profile. Across all four feet and the complete evaluation sequence,
the pooled vertical-force RMSE is $2.728\,\mathrm{N}$. The largest
transient discrepancies occur near touchdown, where the abrupt force
onset and filtering introduce a short response delay.

\begin{figure}[t]
    \vspace{-2mm}
    \centering
    \setlength{\abovecaptionskip}{0pt}
    \includegraphics[width=0.8\columnwidth]
    {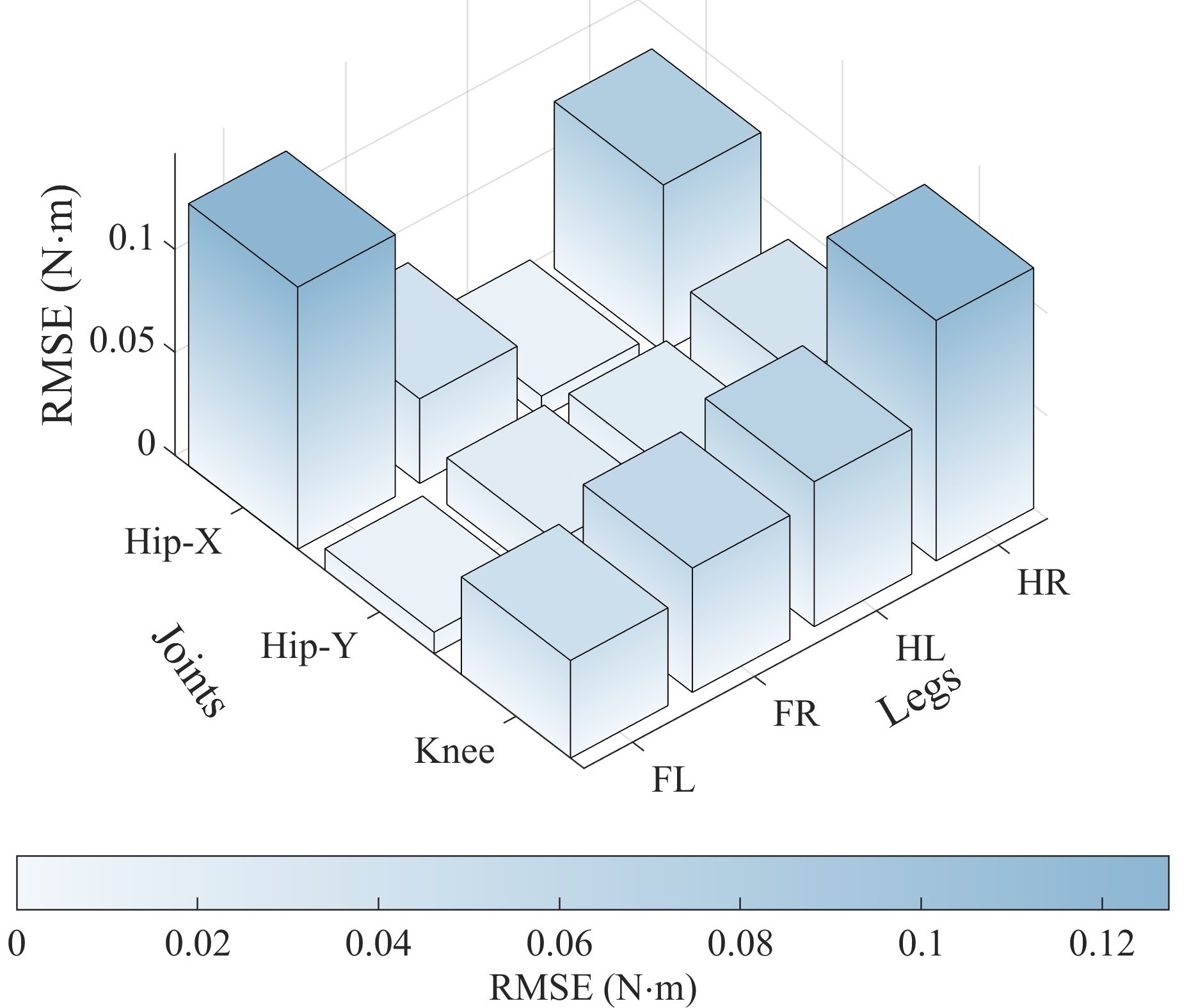}
    \caption{Joint-wise RMSE between DO outputs obtained using estimated
    and MuJoCo contact forces over the complete evaluation sequence.
    The horizontal axes identify the four legs (FL, FR, HL, and HR)
    and the three joint types (Hip-X, Hip-Y, and knee). Each bar represents
    one joint, with its height indicating the RMSE in $\mathrm{N\cdot m}$.}
    \label{fig:disturbance_force_metrics}
\end{figure}

To assess the effect of contact-force substitution on the observer, the same recorded trajectory and observer settings are
used with either the MuJoCo contact forces or the estimated forces as
input. A representative comparison for the HR Hip-X joint is shown in
Fig.~\ref{fig:disturbance_force_curves}, while
Fig.~\ref{fig:disturbance_force_metrics} summarizes the errors across
all twelve joints. The mean joint-wise RMSE is
$0.0542\,\mathrm{N\cdot m}$, and the RMSE pooled over all samples and
joints is $0.0657\,\mathrm{N\cdot m}$. Overall, the force-substitution results show only minor deviations in the observer output under the evaluated conditions, providing empirical support for using sensorless contact-force estimates during deployment. Although model-based force reconstruction remains approximate, the learned policy can respond to residual motion errors through proprioceptive feedback. This integration combines the physical interpretability of model-based estimation with the adaptability of learning, reducing reliance on model accuracy alone.



\subsection{Reinforcement Learning Framework}
\label{sec:rl_framework}

For the RL architecture, we adopt a teacher–student style framework widely used in prior work~\cite{c3,c10,c11,c18} and jointly train the policy, environmental factor encoder, and adaptation module in a single stage. During training, the privileged observation $\mathbf{o}_t^{\mathrm{priv}}\in\mathbb{R}^{54}$ is encoded into the teacher latent $\mathbf{z}_t$, which is provided to the policy together with the current observation $\mathbf{o}_t$ and the disturbance estimation $\hat{\boldsymbol{\tau}}_d$. Meanwhile, the adaptation module infers the student latent $\hat{\mathbf{z}}_t$ from the observation history $\mathbf{o}_t^H$, using $\mathbf{z}_t$ as the regression target. The policy is optimized using proximal policy optimization (PPO)~\cite{c26}, while the adaptation module is trained through latent regression. During deployment, the privileged information and environment factor encoder are removed, and the adaptation module infers $\hat{\mathbf{z}}_t$ from the observation history $\mathbf{o}_t^H$. The policy then uses $\mathbf{o}_t$, $\hat{\mathbf{z}}_t$, and $\hat{\boldsymbol{\tau}}_d$ for action generation.


\subsubsection{Policy Network and Value Network}

As illustrated in Fig.~\ref{fig:framework}, PACE exploits the explicit disturbance estimation $\hat{\boldsymbol{\tau}}_d$ through two complementary pathways.
In addition to torque-level feedforward compensation, $\hat{\boldsymbol{\tau}}_d$ is included in the policy observation to enable disturbance-aware action generation.
During training, the policy is represented as $\pi_{\phi}(\mathbf{a}_t \mid \mathbf{o}_t,\mathbf{z}_t,\hat{\boldsymbol{\tau}}_d)$.
The value network estimates $V_{\psi}(\mathbf{o}_t,\mathbf{z}_t,\hat{\boldsymbol{\tau}}_d)$.
The current observation $\mathbf{o}_t$ consists of the base velocity command, the roll--pitch--yaw
orientation, body angular velocity, current joint positions and velocities, and a short history of
joint states and policy actions. The base velocity command includes the desired linear and yaw velocities. The short history consists of the joint positions $\{\mathbf{q}_{t-i}\}_{i=1}^{3}$, joint velocities $\{\dot{\mathbf{q}}_{t-i}\}_{i=1}^{2}$, and policy actions $\{\mathbf{a}_{t-i}\}_{i=1}^{2}$. The corresponding state measurements are obtained from
onboard joint encoders and the IMU, while the historical quantities are
retained from previous control steps. This short-horizon history provides additional local temporal
information about the robot motion and facilitates the learning of robust locomotion behaviors.



The privileged observation $\mathbf{o}_t^{\mathrm{priv}}$ used by the environment factor encoder contains the
contact states of the four feet, friction coefficient, payload,
CoM shifts, actuator degradation, external force and torque, as well as the proportional and derivative gains of
the PD controller. During training, the privileged observation $\mathbf{o}_t^{\mathrm{priv}}$ is encoded by the environment factor encoder into the teacher latent $\mathbf{z}_t$, which is then provided to both the policy and value networks.

\begin{table}[t]
\centering
\makeatletter
\long\def\@makecaption#1#2{%
    {\footnotesize\noindent #1: #2\par
    \vskip 0.5\baselineskip}}
\makeatother
\caption{Reward structure. $\exp(\cdot)$ is the exponential function, $\lvert\cdot\rvert$ denotes
    absolute value for scalars, and $\lVert\cdot\rVert_2$ denotes the Euclidean norm for vectors.
    The superscripts $(\cdot)^{\mathrm{cmd}}$ and $(\cdot)^{\mathrm{des}}$ indicate commanded and
    desired quantities. $\mathbf{v}_{xy}$, $v_z$, $\boldsymbol{\omega}_{xy}$, and $\omega_z$ are the
    body-frame base linear and angular velocities, $\mathbf{g}_{xy}$ denotes the horizontal
    components of the gravity vector projected into the body frame, and $h$ is the base height.
    $\mathbf{f}_b$ is the contact force on body $b$; $T_k^{\mathrm{air}}$,
    $\mathbf{v}_{\mathrm{f},k}$, and $c_k$ are the air time, velocity, and contact state of foot
    $k$. $\mathbb{I}(\cdot)$ is the indicator function,
    $\mathbb{I}_{\mathrm{cmd}}=\mathbb{I}(\lVert\mathbf{v}_{xy}^{\mathrm{cmd}}\rVert_2>0.1)$, and
    $\mathbb{I}_k^{\mathrm{fc}}$ indicates the first contact of foot $k$. Finally,
    $(x)_+=\max(x,0)$, while $\mathcal{B}_{\mathrm{col}}$ and $\mathcal{J}_{\mathrm{abad}}$ denote
    the penalized bodies and hip abduction--adduction joints.}
\label{tab:reward_function}
\renewcommand{\arraystretch}{1.15}
\setlength{\tabcolsep}{2pt}
\tablebodyfont
\newcommand{\tablemathraise}[1]{\raisebox{1.1pt}[\height][\depth]{#1}}
\begin{tabular}{
    @{}
    >{\raggedright\arraybackslash}m{0.24\columnwidth}
    >{\centering\arraybackslash}m{0.52\columnwidth}
    >{\centering\arraybackslash}m{0.18\columnwidth}
    @{}
}
\hline
\text{Reward} & \text{Equation ($r_i$)} & \text{Weight ($w_i$)} \\
\hline

Lin. velocity tracking
&
$\displaystyle
\exp\!\left(
-4\lVert\mathbf{v}_{xy}^{\mathrm{cmd}}
-\mathbf{v}_{xy}\rVert_2^2
\right)$
&
$1.0$
\\

Yaw velocity tracking
&
$\displaystyle
\exp\!\left(
-4\lvert\omega_z^{\mathrm{cmd}}-\omega_z\rvert^2
\right)$
&
$0.5$
\\

Vertical velocity
&
$\displaystyle v_z^2$
&
$-2.0$
\\

Roll--pitch rate
&
$\displaystyle \lVert\boldsymbol{\omega}_{xy}\rVert_2^2$
&
$-0.05$
\\

Orientation
&
$\displaystyle \lVert\mathbf{g}_{xy}\rVert_2^2$
&
$-0.2$
\\

Base height
&
$\displaystyle (h-h^{\mathrm{des}})^2$
&
$-1.0$
\\

Joint torque
&
$\displaystyle \lVert\boldsymbol{\tau}\rVert_2^2$
&
$-10^{-5}$
\\

Joint acceleration
&
$\displaystyle \lVert\ddot{\mathbf{q}}\rVert_2^2$
&
$-2.5{\times}10^{-7}$
\\

Action rate
&
$\displaystyle
\lVert\mathbf{a}_t-\mathbf{a}_{t-1}\rVert_2^2$
&
$-0.01$
\\

Undesired collision
&
\tablemathraise{%
$\displaystyle
\sum_{b\in\mathcal{B}_{\mathrm{col}}}
\mathbb{I}(\lVert\mathbf{f}_b\rVert_2>0.1)$}%
&
$-1.0$
\\

Joint-position limit
&
\tablemathraise{%
$\displaystyle
\sum_j\!\left[
(q_j^{\min}-q_j)_+
+(q_j-q_j^{\max})_+
\right]$}%
&
$-10.0$
\\

Feet air time
&
\tablemathraise{%
$\displaystyle
\mathbb{I}_{\mathrm{cmd}}
\sum_{k=1}^{4}
\mathbb{I}_k^{\mathrm{fc}}\,
(T_k^{\mathrm{air}}-0.5)$}%
&
$1.0$
\\

Nominal joint posture
&
\tablemathraise{%
$\displaystyle
\sum_{j\in\mathcal{J}_{\mathrm{abad}}}
\lvert q_j-q_j^{\mathrm{default}}\rvert$}%
&
$-0.05$
\\

Foot contact velocity
&
\tablemathraise{%
$\displaystyle
\sum_{k=1}^{4}
c_k\lVert\mathbf{v}_{\mathrm{f},k}\rVert_2$}%
&
$-0.05$
\\

\hline
\end{tabular}
\end{table}

\subsubsection{Action Space}

The policy outputs a 12-dimensional action vector $\mathbf{a}_t \in \mathbb{R}^{12}$, representing
joint-position offsets for the robot's twelve actuated joints. Rather than directly predicting the
absolute desired joint positions, the policy generates these offsets with respect to a predefined
nominal joint-position vector $\mathbf{q}_{\mathrm{stand}}\in\mathbb{R}^{12}$, which represents the robot's standing
configuration. Accordingly, the desired joint-position vector is computed as
\begin{equation}
\label{eq:action_space}
\mathbf{q}_{\mathrm{des}}
=
\mathbf{q}_{\mathrm{stand}}
+
\mathbf{a}_t,
\end{equation}
The resulting desired joint positions $\mathbf{q}_{\mathrm{des}}$ are subsequently tracked by the
joint-level PD controller in Eq.~\eqref{eq:pd_control}, whose torque output is combined with disturbance feedforward compensation according to Eq.~\eqref{eq:do_compensation}.

\subsubsection{Reward Function}

To avoid attributing the performance improvement to extensive reward engineering, our reward
function follows formulations commonly used in previous studies of quadrupedal
locomotion~\cite{c10,c11,c12,c20,c27}. As summarized in Table~\ref{tab:reward_function}, the reward
terms encourage accurate velocity tracking, stable body motion, smooth joint behavior, and
appropriate foot contacts. The total reward at time step $t$ is defined as
\begin{equation}
\label{eq:reward}
r_t = \sum_i w_i r_i ,
\end{equation}
where $r_i$ and $w_i$ denote the individual reward term and its corresponding weight, respectively.
No additional reward term is introduced specifically for the DO, allowing the
effect of the proposed PACE framework to be evaluated without relying on task-specific reward
redesign.

\section{EXPERIMENTS}

\subsection{Experiment Setup}

The proposed method is evaluated on a legged robot named DeepRobotics Lite3. We employed the
Isaac Gym simulator~\cite{c29} with the legged gym framework~\cite{c12} to train the locomotion
policies for 15,000 iterations. During training, 4,096 parallel environments with domain
randomization are executed simultaneously. The ranges of randomized physical parameters are
summarized in Table~\ref{tab:domain_randomization}. All compared methods are optimized using the PPO
algorithm with the same hyperparameters, including a clipping coefficient of 0.2, a generalized
advantage estimation factor of 0.95, and a discount factor of 0.99. The network parameters are
updated by the \textit{Adam} optimizer~\cite{c30} with a learning rate of $10^{-3}$. All methods are trained using the same reward functions described in Table~\ref{tab:reward_function}. 

For a fair comparison, all methods use MLPs as their basic network architecture, the exponential linear unit (ELU) activation function~\cite{c28}, and fixed random seeds. Moreover, all compared policies are trained exclusively on flat terrain, without exposure to the height discontinuities or challenging terrains. These unseen conditions are not covered by the domain randomization used during training. Nevertheless, PACE maintains robust locomotion under such conditions, as demonstrated in the subsequent real-world experiments.

\begin{table}[t]
\centering
\makeatletter
\long\def\@makecaption#1#2{%
    {\footnotesize\noindent #1: #2\par
    \vskip 0.5\baselineskip}}
\makeatother
\caption{Domain randomization ranges applied in the simulation.}
\label{tab:domain_randomization}
\renewcommand{\arraystretch}{1.15}
\setlength{\tabcolsep}{2pt}
\tablebodyfont
\begin{tabular}{
    @{}
    >{\raggedright\arraybackslash}p{0.42\columnwidth}
    >{\centering\arraybackslash}p{0.34\columnwidth}
    >{\centering\arraybackslash}p{0.18\columnwidth}
    @{}
}
\hline
\text{Parameter} & \text{Randomization range} & \text{Unit} \\
\hline

Added base mass
& $[-1,\,3]$
& $\mathrm{kg}$ \\

$K_{\mathrm{p}}$ factor
& $[0.8,\,1.2]$
& -- \\

$K_{\mathrm{d}}$ factor
& $[0.8,\,1.2]$
& -- \\

Motor strength factor
& $[0.8,\,1.2]$
& -- \\

CoM shift ($x$)
& $[-50,\,10]$
& $\mathrm{mm}$ \\

CoM shift ($y$)
& $[-30,\,30]$
& $\mathrm{mm}$ \\

CoM shift ($z$)
& $[-30,\,30]$
& $\mathrm{mm}$ \\

Friction coefficient
& $[0.1,\,1.25]$
& -- \\

\hline
\end{tabular}
\end{table}

The policy training is conducted on a workstation equipped with an NVIDIA GeForce RTX 3060 GPU and
an Intel Xeon E5-2683 v4 CPU. In the real-world experiments, the proposed framework is directly
deployed on the Lite3 robot. During real-world deployment, the DO is updated with a sampling period of $0.003~\mathrm{s}$, enabling real-time disturbance estimation.

\subsection{Compared Methods}

For a comparative evaluation, we compare the proposed method with several representative
RL-based locomotion frameworks, as specified below.

\begin{itemize}
    \item \textbf{BASELINE}~\cite{deep}: The policy was trained with an implicit environment factor
        encoder in a teacher--student style framework.

    \item \textbf{HIMLOCO}~\cite{c21}: The policy was trained with a hierarchical framework that
        infers environmental information from the internal response of the robot.

    \item \textbf{PACE$_0$}: The policy was enhanced by feedforward compensation using explicit disturbance estimation, while the estimation was excluded from the policy observation. This feedforward-only configuration was retained at deployment.

    \item \textbf{PACE}: The policy was enhanced by feedforward compensation using explicit disturbance estimation and further augmented with the estimates as additional observations. Both pathways were retained at deployment.
\end{itemize}

All four methods are evaluated in simulation. In experiments on the physical robot, BASELINE, PACE$_0$ and PACE are evaluated.

\subsection{Experiment Results}

We evaluate the proposed framework under persistent dynamics variations, out-of-distribution (OOD) commands, transient external disturbances, and zero-shot real-world conditions. Together, these tests assess both tracking performance and robustness to model mismatch, external perturbations, and unseen operating conditions.

\subsubsection{Command Tracking under Dynamics Variations}

To evaluate robustness against unknown dynamics variations, we test all controllers in simulation
under additional payloads, CoM shifts, and actuator degradation. The
commanded horizontal linear velocity and yaw rate are fixed at
$(v_x^{\mathrm{cmd}},v_y^{\mathrm{cmd}}) =(1,0)~\mathrm{m/s}$ and
$\omega_z^{\mathrm{cmd}}=0~\mathrm{rad/s}$, respectively. Tracking performance is measured by
$\mathrm{RMSE}_{xy} =\sqrt{\frac{1}{N}\sum_{i=1}^{N} [(v_x^{i}-1)^2+(v_y^{i})^2]}$, where $N$ is the
number of samples.

As summarized in Table~\ref{tab:dynamics_variation}, PACE achieves the lowest
$\mathrm{RMSE}_{xy}$ in all tested conditions. Averaged over the eleven conditions, it reduces the
tracking error by $57.0\%$, $18.1\%$, and $32.8\%$ compared with BASELINE, HIMLOCO, and PACE$_0$,
respectively. Moreover, PACE$_0$ consistently achieves lower tracking errors than BASELINE, indicating improved performance with feedforward compensation. Building on PACE$_0$, the full PACE framework further incorporates disturbance estimation into the policy observations. It achieves the lowest tracking errors across all tested conditions.

\begin{table}[t]
\centering
\makeatletter
\long\def\@makecaption#1#2{%
    {\footnotesize\noindent #1: #2\par
    \vskip 0.5\baselineskip}}
\makeatother
\caption{Command-tracking performance under dynamics variations. All
    entries are $\mathrm{RMSE}_{xy}$ values in $\mathrm{m/s}$, and lower values indicate better
    performance.}
\label{tab:dynamics_variation}

\renewcommand{\arraystretch}{1.12}
\setlength{\tabcolsep}{1.0pt}
\tablebodyfont

\begin{tabular}{
    @{}
    >{\raggedright\arraybackslash}m{0.30\columnwidth}
    >{\centering\arraybackslash}m{0.17\columnwidth}
    >{\centering\arraybackslash}m{0.17\columnwidth}
    >{\centering\arraybackslash}m{0.14\columnwidth}
    >{\centering\arraybackslash}m{0.14\columnwidth}
    @{}
}
\hline
\noalign{\vskip 1.5pt}

Condition
& BASELINE
& HIMLOCO
& PACE$_0$
& PACE \\

\hline

Payload: $5~\mathrm{kg}$
& 0.398 & 0.193 & 0.252 & \textbf{0.162} \\

Payload: $8~\mathrm{kg}$
& 0.443 & 0.263 & 0.262 & \textbf{0.174} \\

Payload: $11~\mathrm{kg}$
& 0.476 & 0.246 & 0.279 & \textbf{0.194} \\

\noalign{\vskip 2pt}

CoM shift: $x=2~\mathrm{cm}$
& 0.369 & 0.200 & 0.238 & \textbf{0.169} \\

CoM shift: $y=3~\mathrm{cm}$
& 0.370 & 0.201 & 0.237 & \textbf{0.169} \\

CoM shift:
$x=2,\ y=3~\mathrm{cm}$
& 0.365 & 0.197 & 0.235 & \textbf{0.170} \\

\noalign{\vskip 2pt}

All motors weakened to $80\%$
& 0.353 & 0.176 & 0.235 & \textbf{0.157} \\

Left motors weakened to $80\%$
& 0.339 & 0.184 & 0.233 & \textbf{0.145} \\

Right motors weakened to $80\%$
& 0.347 & 0.166 & 0.225 & \textbf{0.148} \\

Front motors weakened to $80\%$
& 0.336 & 0.160 & 0.226 & \textbf{0.145} \\

Rear motors weakened to $80\%$
& 0.354 & 0.190 & 0.233 & \textbf{0.150} \\

\hline
\end{tabular}
\end{table}

\subsubsection{Command Tracking under OOD Commands}

To evaluate generalization beyond the training command distribution,
we test the learned policies under an OOD
velocity command. During training, commanded linear velocities are
uniformly sampled from $[-1,1]\,\mathrm{m/s}$. During evaluation,
the commands are set to $v_x^{\mathrm{cmd}}=2\,\mathrm{m/s}$,
$v_y^{\mathrm{cmd}}=0\,\mathrm{m/s}$, and
$\omega_z^{\mathrm{cmd}}=0\,\mathrm{rad/s}$.
Tracking performance is measured using the absolute velocity errors
$v_x^e=|v_x-v_x^{\mathrm{cmd}}|$ and
$v_y^e=|v_y-v_y^{\mathrm{cmd}}|$.

\begin{figure}[t]
    \centering
    \setlength{\abovecaptionskip}{0pt}
    \includegraphics[width=\columnwidth]
    {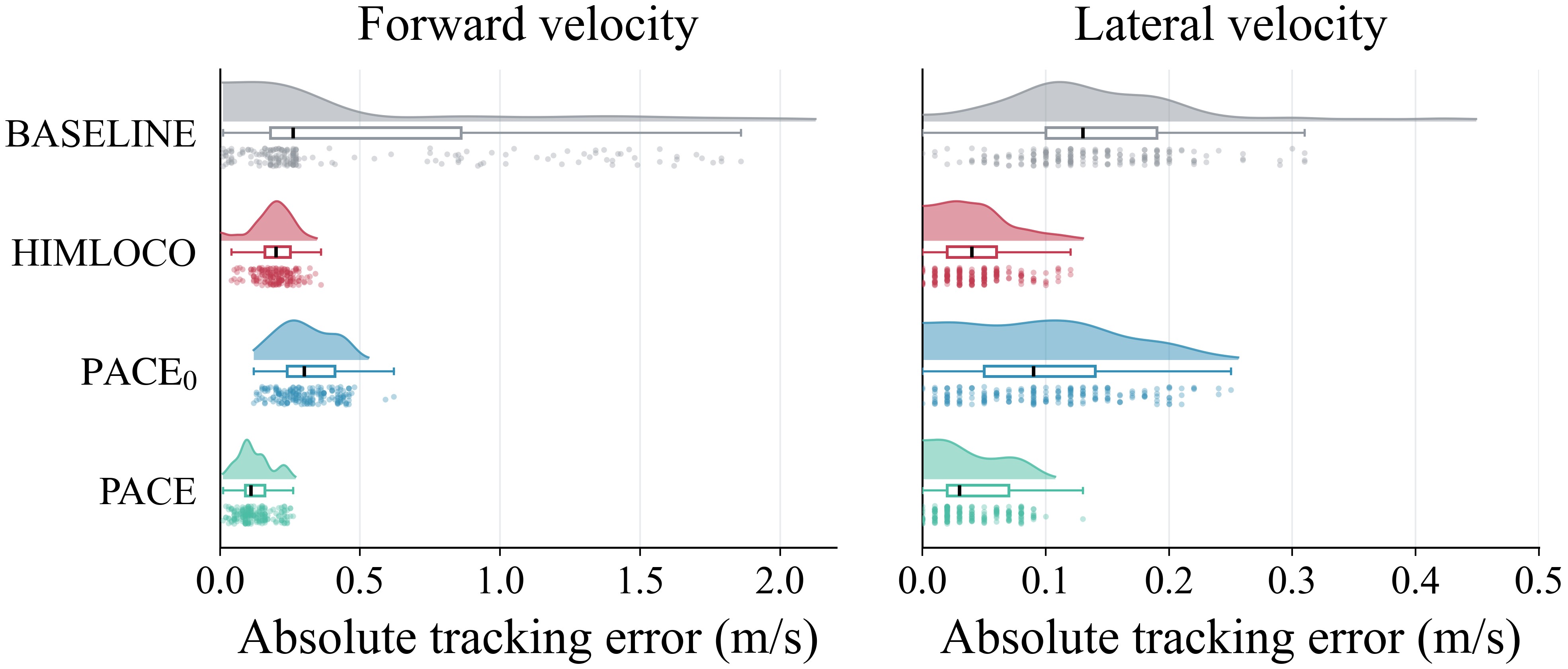}

\caption{Raincloud distributions of absolute forward (left) and
lateral (right) velocity tracking errors under the OOD 
command of $2\,\mathrm{m/s}$. Half-violins show density estimates
scaled to equal peak height. Boxes indicate the interquartile
range (IQR), with internal lines marking the median, and whiskers
extend to the most extreme samples within the $1.5\,\mathrm{IQR}$
fences. Dots represent the displayed time samples.}
    \label{fig:ood_tracking}
\end{figure}

Fig.~\ref{fig:ood_tracking} presents raincloud distributions of the absolute velocity tracking errors over the complete $0$--$12\,\mathrm{s}$ trajectories, including the initial transient. All samples are retained for
density estimation and descriptive analysis. For display, each
cloud ends at the rightmost downward crossing of $5\%$ of its
peak density.

BASELINE initially falls repeatedly under the unseen command, resulting in large tracking errors that are retained despite automatic resets. Compared with BASELINE, PACE$_0$ reduces the mean forward and lateral
errors from $0.565$ and $0.162\,\mathrm{m/s}$ to $0.374$ and
$0.101\,\mathrm{m/s}$, respectively. With disturbance estimation additionally
provided to the policy, PACE achieves the lowest median errors
among the four methods: $0.11\,\mathrm{m/s}$ in the forward direction
and $0.03\,\mathrm{m/s}$ in the lateral direction. Its corresponding
95th-percentile errors are also the lowest, at $0.25$ and
$0.09\,\mathrm{m/s}$, indicating reduced upper-tail errors.
These results show improved OOD command tracking with feedforward compensation and support the additional benefit of observation augmentation in the full PACE framework.

\subsubsection{Recovery Capability against External Pushes}

The previous experiments evaluate robustness under persistent
dynamics variations and unseen velocity commands. To further assess
recovery from short-duration external disturbances, we conduct
lateral push experiments in simulation.

\begin{figure}[!t]
    \centering
    \setlength{\abovecaptionskip}{0pt}
    \includegraphics[width=\columnwidth]
    {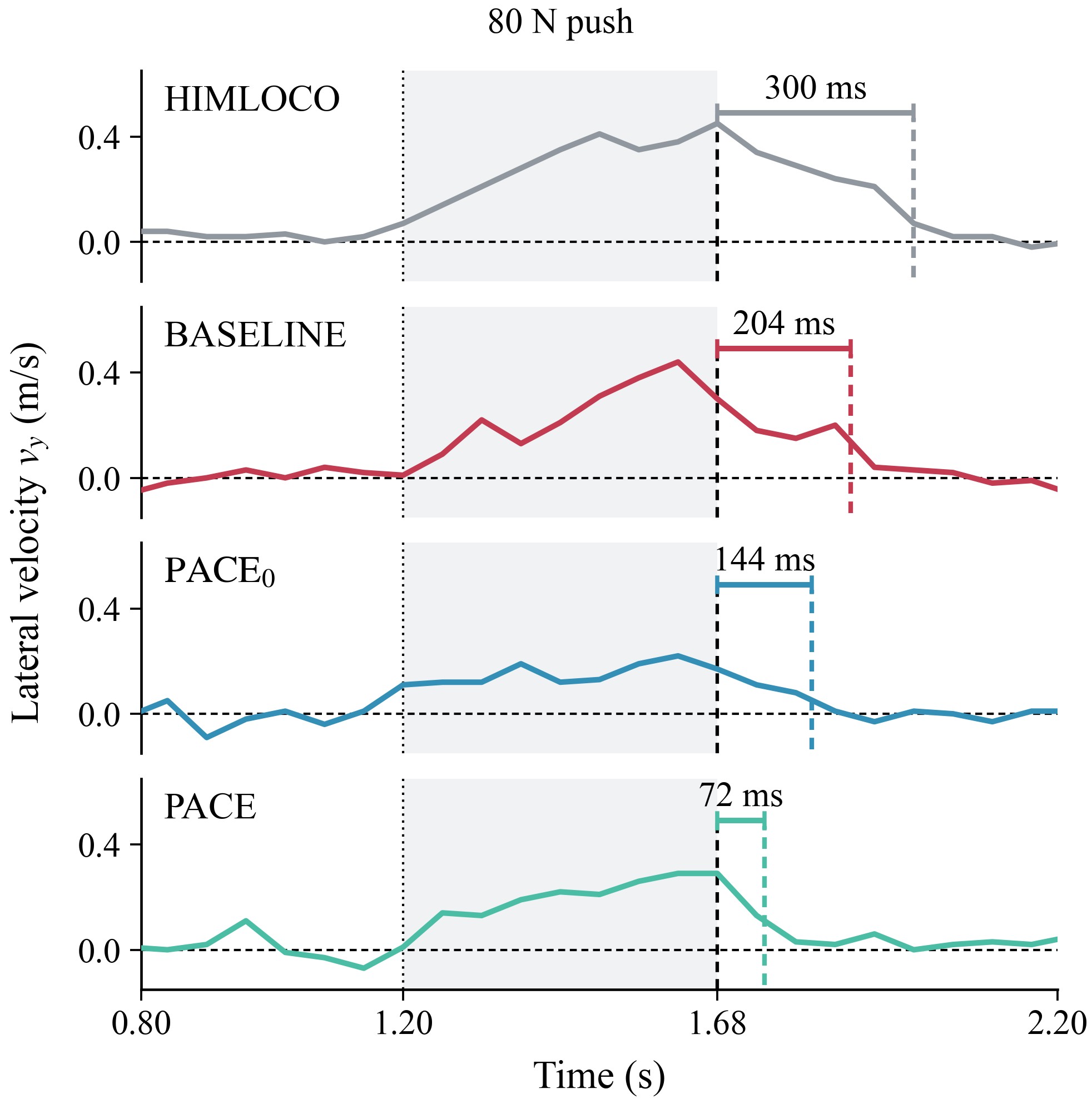}
    \caption{Lateral velocity responses to an $80\,\mathrm{N}$ push
    along the $y$ direction during forward locomotion commanded at
    $1\,\mathrm{m/s}$. Shaded regions indicate force application from
    $1.20$ to $1.68\,\mathrm{s}$. Colored vertical dashed lines mark
    the recovery instants, which are determined using the full planar velocity and yaw-rate criteria defined in the text.}
    \label{fig:push_recovery}
\end{figure}

During the test, the commanded velocities are fixed at
$v_x^{\mathrm{cmd}}=1\,\mathrm{m/s}$,
$v_y^{\mathrm{cmd}}=0\,\mathrm{m/s}$, and
$\omega_z^{\mathrm{cmd}}=0\,\mathrm{rad/s}$.
An $80\,\mathrm{N}$ external force is applied along the lateral
($y$) direction at $1.20\,\mathrm{s}$ and maintained for
$0.48\,\mathrm{s}$, before being removed at $1.68\,\mathrm{s}$.
Fig.~\ref{fig:push_recovery} presents the resulting lateral velocity responses.
The robot is considered recovered when the Euclidean norm of the
planar velocity tracking error is below $0.12\,\mathrm{m/s}$ and
the absolute yaw-rate tracking error is below
$0.12\,\mathrm{rad/s}$, with both conditions maintained for
$0.12\,\mathrm{s}$. Recovery time is measured from force removal
to the beginning of the first interval satisfying this criterion,
preventing a brief threshold crossing from being counted as recovery.

The recovery times of HIMLOCO, BASELINE, PACE$_0$, and PACE are
$300$, $204$, $144$, and $72\,\mathrm{ms}$, respectively.
Both PACE variants exhibit smaller peak lateral velocity deviations
and faster recovery than the baselines. PACE achieves the
shortest recovery time, corresponding to reductions of $64.7\%$
and $76.0\%$ relative to BASELINE and HIMLOCO, respectively.
This rapid recovery reflects the combined closed-loop response of the learned policy and disturbance compensation, rather than the convergence speed of the observer alone. PACE incorporates both feedforward compensation and observation augmentation during training and retains these pathways at deployment. This preserves the disturbance-aware control structure in which the policy was trained. Such consistency contributes to the rapid recovery observed in the external-push experiments.

\begin{figure*}[!t]
    \centering
    \setlength{\abovecaptionskip}{0pt}
    \includegraphics[width=0.8\textwidth]{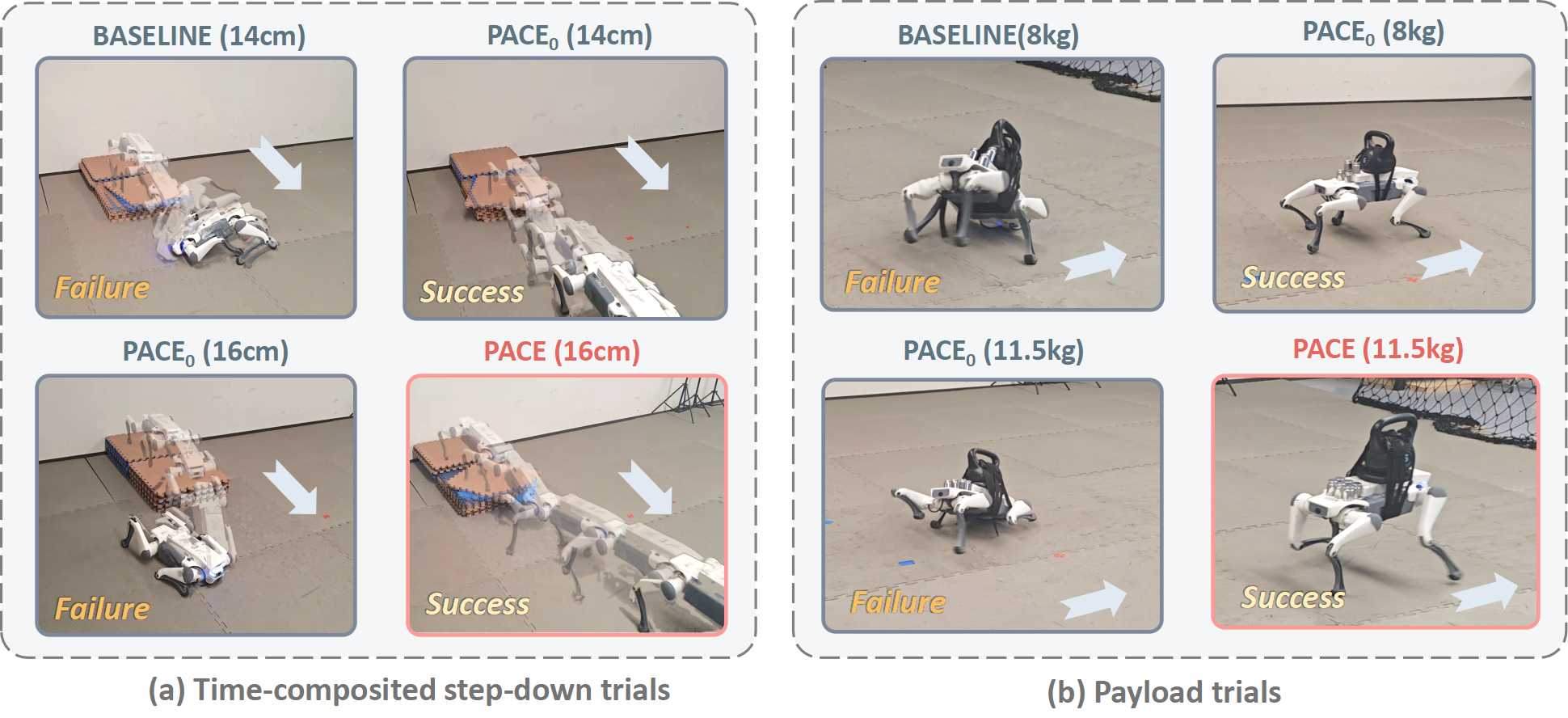}
    \caption{Real-world robustness experiments.
    (a) Time-composited step-down trials: BASELINE fails at
    $14~\mathrm{cm}$, whereas PACE$_0$ successfully recovers;
    at $16~\mathrm{cm}$, PACE$_0$ fails while PACE maintains
    stable locomotion.
    (b) Payload trials: BASELINE fails with an $8~\mathrm{kg}$
    payload, whereas PACE$_0$ successfully traverses under the same load; with an
    $11.5~\mathrm{kg}$ payload, PACE$_0$ fails while PACE
    maintains stable forward locomotion.
    Successive overlaid poses in (a) indicate temporal
    progression, and white arrows denote the commanded
    travel direction.}
    \label{fig:real_world_experiments}
\end{figure*}


\subsubsection{Maximum Step-Down Capability}

To evaluate zero-shot robustness to unseen height discontinuities, we conduct real-world step-down
experiments. The reported platform height is varied from $8~\mathrm{cm}$ to $16~\mathrm{cm}$ in increments of
$2~\mathrm{cm}$, and each method is tested five times at every height. A trial is considered
successful if the robot descends without falling and subsequently maintains stable forward
locomotion. This test evaluates both zero-shot generalization to unseen terrain discontinuities and robustness
to the transient contact disturbances induced during landing. 

As shown in Table~\ref{tab:step_down},
PACE achieves the highest overall success rate and completes all trials at $16~\mathrm{cm}$,
 the maximum successful step-down height observed in our experiments. Additional tests at greater heights were unsuccessful for all methods. Representative motion sequences are shown in Fig.~\ref{fig:real_world_experiments}(a). At
$14~\mathrm{cm}$, BASELINE fails after landing whereas PACE$_0$ successfully recovers. At
$16~\mathrm{cm}$, PACE$_0$ fails while PACE remains stable and continues locomotion. These
results demonstrate that the proposed framework substantially improves real-world step-down
robustness.

\subsubsection{Maximum Payload Capacity}

To evaluate zero-shot robustness to large unseen payload variations, we conduct real-world payload tests. The payload attached to the robot base is progressively increased until the robot can no longer complete the prescribed locomotion trial without falling. 

In the payload tests, BASELINE maintains stable locomotion with a maximum payload of $7.5~\mathrm{kg}$ and fails at $8~\mathrm{kg}$. In comparison, PACE$_0$ increases this capacity to $11~\mathrm{kg}$, while failing at $11.5~\mathrm{kg}$. PACE further extends the payload capacity to $11.5~\mathrm{kg}$ while maintaining stable forward locomotion, which is nearly equal to the robot's own mass of $11.6025~\mathrm{kg}$.  Tests with greater payloads were unsuccessful. Representative trials are shown in Fig.~\ref{fig:real_world_experiments}(b). These results demonstrate that the proposed framework substantially extends the payload operating range of the locomotion policy, providing direct real-world evidence of improved robustness to large dynamics variations.

\subsubsection{Unseen Terrain Test}

Since all policies are trained exclusively on flat terrain, we further evaluate their zero-shot
generalization on five unseen real-world terrains, including flagstone, lawn, sand, gravel road, and
large rocks. Each method is evaluated over five traversal trials on every terrain. A trial is
considered successful if the robot traverses the test segment without falling. The corresponding
success rates are summarized in Table~\ref{tab:unseen_terrain}.

As shown in Table~\ref{tab:unseen_terrain}, both PACE variants substantially outperform BASELINE across
the unseen terrains, demonstrating improved terrain adaptability and zero-shot generalization.
Representative real-world terrain traversals are shown in Fig.~\ref{fig:framework}.
Although PACE$_0$ achieves a slightly higher success rate than PACE on the large rocks, both
variants remain considerably more robust than BASELINE. This result suggests that the benefit from observation augmentation can vary under highly irregular contact conditions.

\begin{table}[t]
\centering
\makeatletter
\long\def\@makecaption#1#2{%
    {\footnotesize\noindent #1: #2\par
    \vskip 0.5\baselineskip}}
\makeatother
\caption{Step-down success rates on unseen height discontinuities. All entries are success rates
    (\%), and each method is evaluated over five trials at each height.}
\label{tab:step_down}
\renewcommand{\arraystretch}{1.12}
\setlength{\tabcolsep}{5pt}

\tablebodyfont
\begin{tabular}{lccccc}
\hline
Method
& 8 cm
& 10 cm
& 12 cm
& 14 cm
& 16 cm
\\
\hline

BASELINE
& 80
& 0
& 0
& 0
& 0
\\

PACE$_0$
& \textbf{100}
& \textbf{100}
& \textbf{100}
& 80
& 0
\\

PACE
& \textbf{100}
& \textbf{100}
& \textbf{100}
& \textbf{100}
& \textbf{100}
\\

\hline
\end{tabular}
\end{table}

\begin{table}[t]
\centering
\makeatletter
\long\def\@makecaption#1#2{%
    {\footnotesize\noindent #1: #2\par
    \vskip 0.5\baselineskip}}
\makeatother
\caption{Traversal success rates on unseen real-world terrains. All entries are success rates (\%),
    and each method is evaluated over five trials on each terrain.}
\label{tab:unseen_terrain}
\renewcommand{\arraystretch}{1.12}
\setlength{\tabcolsep}{3.5pt}

\tablebodyfont
\begin{tabular}{lccccc}
\hline
Method
& \shortstack{Flagstone}
& Lawn
& Sand
& Gravel road
& \shortstack{Large rocks}
\\
\hline

BASELINE
& 60
& 0
& 0
& 40
& 0
\\

PACE$_0$
& \textbf{100}
& \textbf{100}
& \textbf{100}
& \textbf{100}
& \textbf{80}
\\

PACE
& \textbf{100}
& \textbf{100}
& \textbf{100}
& \textbf{100}
& 60
\\

\hline
\end{tabular}
\end{table}

The results show that PACE remains effective under substantial dynamics variations, external disturbances, and unseen real-world conditions despite the deliberately simplified nominal dynamics used by the DO. This supports the robustness of the lumped-disturbance formulation to model uncertainty and demonstrates its practical applicability on the real robot.

\section{CONCLUSIONS}


We presented PACE, a locomotion framework that lumps model uncertainties and external effects into a single joint-space disturbance. An acceleration-free nonlinear observer estimates this disturbance online. The observer remains in the closed loop throughout both training and deployment. The estimation provides foundational torque-level feedforward compensation, while its additional incorporation into the policy observation enables disturbance-aware action generation. A sensorless foot-force estimator supplies the contact-force term from onboard measurements, so the same observer structure runs in simulation and on hardware. In simulation, PACE lowers the velocity-tracking error by $57.0\%$ and $18.1\%$ over BASELINE and HIMLOCO, and on a Lite3 trained only on flat ground, it walks under an $11.5~\mathrm{kg}$ payload and clears $16~\mathrm{cm}$ step-downs.

\begin{figure*}[!t]
    \centering
    \includegraphics[width=\textwidth]{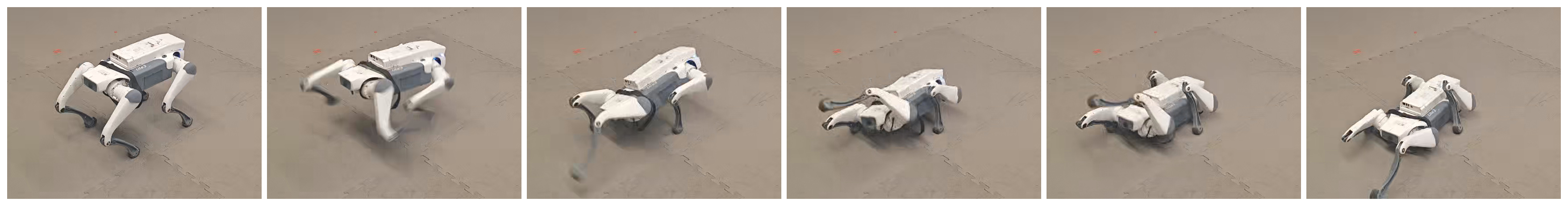}
    \caption{Failure case of direct acceleration-based disturbance estimation on hardware. 
    In this variant, the disturbance used for feedforward compensation and policy augmentation 
    is computed directly from reconstructed acceleration-related quantities, instead of being 
    estimated by the proposed DO. The successive frames, ordered from left to right, 
    show that the Lite3 rapidly loses stability and falls during deployment. This result indicates 
    that directly using reconstructed acceleration information to compute the disturbance is highly 
    sensitive to real-world sensing noise and reconstruction errors, whereas the proposed observer-based 
    formulation provides a more reliable disturbance signal for deployment.}
    \label{fig:acc_direct_failure}
\end{figure*}

A limitation of the current formulation is that the DO uses a fixed scalar gain $L$ for all joints and operating conditions. Although this setting performs effectively across the evaluated scenarios, an adaptive or joint-dependent observer gain could better balance disturbance-tracking speed and sensitivity to estimation noise. Future work will investigate such adaptive observer designs and extend PACE to humanoid platforms and contact-rich manipulation.









\appendix
\phantomsection
\section{Implementation Details of the Sensorless Foot-Force Estimator}
\label{app:preprocessing}
This appendix provides the signal preprocessing used to obtain the
joint acceleration and floating-base motion quantities required by the
recursive Newton--Euler formulation in Section~IV-C. At control step
$k$, the estimator uses the measured joint velocity
$\dot{\mathbf{q}}_k$, body angular velocity
$\boldsymbol{\omega}_{\mathrm{B},k}$, specific force
$\mathbf{f}_{\mathrm{B},k}$, and body-to-world rotation matrix
$\mathbf{R}_{\mathrm{WB},k}$.

The joint acceleration and base angular acceleration are first
approximated by finite differences,
\begin{equation}
\label{eq:ffe_derivative_app}
\ddot{\mathbf{q}}_{k}^{\mathrm{raw}}
=
\frac{\dot{\mathbf{q}}_{k}-\dot{\mathbf{q}}_{k-1}}{\Delta t},
\qquad
\boldsymbol{\alpha}_{\mathrm{B},k}^{\mathrm{raw}}
=
\frac{
\boldsymbol{\omega}_{\mathrm{B},k}
-
\boldsymbol{\omega}_{\mathrm{B},k-1}
}
{\Delta t},
\end{equation}
where $\Delta t$ denotes the control period,
$\ddot{\mathbf{q}}_{k}^{\mathrm{raw}}\in\mathbb{R}^{12}$
is the finite-difference joint acceleration, and
$\boldsymbol{\alpha}_{\mathrm{B},k}^{\mathrm{raw}}\in\mathbb{R}^{3}$
is the finite-difference base angular acceleration.

To suppress numerical differentiation and measurement noise, the
quantities used in inverse dynamics are filtered according to
\begin{equation}
\label{eq:ffe_pre_filter_app}
\begin{aligned}
\bar{\ddot{\mathbf{q}}}_{k}
&=
\alpha_{\mathrm{d}}\ddot{\mathbf{q}}_{k}^{\mathrm{raw}}
+
(1-\alpha_{\mathrm{d}})
\bar{\ddot{\mathbf{q}}}_{k-1},\\
\bar{\boldsymbol{\alpha}}_{\mathrm{B},k}
&=
\alpha_{\mathrm{d}}
\boldsymbol{\alpha}_{\mathrm{B},k}^{\mathrm{raw}}
+
(1-\alpha_{\mathrm{d}})
\bar{\boldsymbol{\alpha}}_{\mathrm{B},k-1},\\
\bar{\mathbf{f}}_{\mathrm{B},k}
&=
\alpha_{\mathrm{d}}\mathbf{f}_{\mathrm{B},k}
+
(1-\alpha_{\mathrm{d}})
\bar{\mathbf{f}}_{\mathrm{B},k-1},
\end{aligned}
\end{equation}
where
$\bar{\ddot{\mathbf{q}}}_{k}\in\mathbb{R}^{12}$
and
$\bar{\boldsymbol{\alpha}}_{\mathrm{B},k},
\bar{\mathbf{f}}_{\mathrm{B},k}\in\mathbb{R}^{3}$
are the filtered signals, and $\alpha_d$ is the preprocessing low-pass
filtering coefficient. In this work, $\alpha_{\mathrm{d}}=0.45$.

The filtered IMU quantities are then transformed into the world frame
to initialize the floating-base recursive dynamics,
\begin{equation}
\label{eq:ffe_base_motion_app}
\begin{aligned}
\boldsymbol{\omega}_{0}^{\mathrm{W}}
&=
\mathbf{R}_{\mathrm{WB}}
\boldsymbol{\omega}_{\mathrm{B}},
\qquad
\boldsymbol{\alpha}_{0}^{\mathrm{W}}
=
\mathbf{R}_{\mathrm{WB}}
\bar{\boldsymbol{\alpha}}_{\mathrm{B}},
\\
\mathbf{a}_{0}^{\mathrm{W}}
&=
\mathbf{R}_{\mathrm{WB}}
\bar{\mathbf{f}}_{\mathrm{B}}
+
\mathbf{g}^{\mathrm{W}},
\end{aligned}
\end{equation}
where
$\boldsymbol{\omega}_{0}^{\mathrm{W}}, 
\boldsymbol{\alpha}_{0}^{\mathrm{W}}, 
\mathbf{a}_{0}^{\mathrm{W}}\in\mathbb{R}^{3}$ denote the base angular velocity,
angular acceleration, and linear acceleration expressed in the world
frame, respectively. The gravitational acceleration is
\begin{equation}
\mathbf{g}^{\mathrm{W}}
=
[0,0,-9.81]^{\mathbf{T}}
~\mathrm{m/s^2}.
\end{equation}

The filtered joint acceleration
$\bar{\ddot{\mathbf{q}}}_{k}$, the transformed floating-base motion
quantities
$\{\boldsymbol{\omega}_{0}^{\mathrm{W}},
\boldsymbol{\alpha}_{0}^{\mathrm{W}},
\mathbf{a}_{0}^{\mathrm{W}}\}$,
the joint states, and the known inertial parameters are next 
used by the recursive \textit{Newton-Euler} formulation described in
Section~IV-C.

Although reconstructed acceleration information is available
for inverse dynamics, it is not directly used for disturbance
estimation in PACE. Instead, it serves as an intermediate
quantity for foot-force reconstruction, where filtering,
regularization, and physical constraints help mitigate
estimation errors. The resulting contact forces are then supplied to the
acceleration-free DO, which provides
an additional layer of disturbance suppression to mitigate
the influence of measurement and reconstruction errors.

To examine this alternative, we trained a policy using
disturbance estimation directly computed from reconstructed
accelerations via inverse dynamics. The estimation replaced
the observer outputs in both feedforward compensation and
observation augmentation. During real-world deployment, this variant rapidly
lost stability and fell, as shown in
Fig.~\ref{fig:acc_direct_failure}. 
This failure suggests that direct acceleration-based
disturbance estimation is less robust to sensing and
reconstruction errors under the tested hardware conditions.
It also supports the observer-based design adopted in PACE.

\bibliographystyle{IEEEtran}
\bibliography{references}

\end{document}